# Explore the Use of Time Series Foundation Model for Car-Following Behavior Analysis

**Luwei Zeng**
Post-Doctoral Researcher
Department of Civil, Environmental & Architectural Engineering
University of Kansas, Lawrence, KS 66045
luweizeng@ku.edu

**Runze Yan**
Post-Doctoral Researcher
Center for Data Science
Nell Hodgson Woodruff School of Nursing
Emory University, Atlanta, GA, 30322
runze.yan@emory.edu

## ABSTRACT

Modeling car-following behavior is essential for traffic simulation, analyzing driving patterns, and understanding complex traffic flows with varying levels of autonomous vehicles. Traditional models like the Safe Distance Model and Intelligent Driver Model (IDM) require precise parameter calibration and often lack generality due to simplified assumptions about driver behavior. While machine learning and deep learning methods capture complex patterns, they require large labeled datasets. Foundation models provide a more efficient alternative. Pre-trained on vast, diverse time series datasets, they can be applied directly to various tasks without the need for extensive re-training. These models generalize well across domains, and with minimal fine-tuning, they can be adapted to specific tasks like car-following behavior prediction. In this paper, we apply Chronos, a state-of-the-art public time series foundation model, to analyze car-following behavior using the Open ACC dataset. Without fine-tuning, Chronos outperforms traditional models like IDM and Exponential smoothing with trend and seasonality (ETS), and achieves similar results to deep learning models such as DeepAR and TFT, with an RMSE of 0.60. After fine-tuning, Chronos reduces the error to an RMSE of 0.53, representing a 33.75% improvement over IDM and a 12–37% reduction compared to machine learning models like ETS and deep learning models including DeepAR, WaveNet, and TFT. This demonstrates the potential of foundation models to significantly advance transportation research, offering a scalable, adaptable, and highly accurate approach to predicting and simulating car-following behaviors.

## INTRODUCTION

Car-following models have been studied for decades using diverse approaches. Modeling car-following behaviors is crucial for microscopic traffic simulation and the analysis of human driving behaviors, and it is also important to understanding the complex traffic flow with various levels of autonomous vehicles. As noted in review articles (Van Wageningen-Kessels et al., 2015; Wang et al., 2023), a considerable amount of mathematical car-following models has been developed, such as safe-distance model (Newell, 1961) and Intelligent Driver Model (IDM) (Treiber et al., 2000), and some new models are built upon them (Milanés & Shladover, 2014; Schakel et al., 2010; Yang et al., 2023). However, these car-following models depend on mathematical formulas, requiring meticulous parameter calibration before traffic simulation—a crucial but challenging process (Li et al., 2016; Ma & Qu, 2020). Such models may lack generalizability, as their parameters might perform poorly with different datasets due to oversimplified assumptions about driver behavior and neglect of individual driving differences and diverse driving styles (Chen et al., 2023). With the recent advancements in machine learning technologies, there has been a shift towards using these methods to directly extract car-following behavior from large field data sets (Ma & Qu, 2020). This has led to the development of data-driven models (Chen et al., 2023). For example, as one of the prevailing machine learning methods, Long Short-Term Memory (LSTM) and Sequence to Sequence (Seq2Seq) has been intensively applied in car-following behavior modeling (Fan et al., 2019; Huang et al., 2018; Lin et al., 2022; Ma & Qu, 2020; Morton et al., 2017). Despite their flexibility and ability to capture complex driving patterns, have limitations. They require substantial amounts of labeled data, can be computationally intensive, and often face challenges with generalization, potentially overfitting to specific datasets and performing poorly on unseen data (Zhang et al., 2021).

Foundation models are advanced machine learning models pre-trained on extensive and diverse datasets. These models are designed to be highly versatile and adaptable, with public parameters, often numbering in the billions, that allow them to be fine-tuned for a wide range of downstream tasks with minimal re-training (Zhao et al., 2024). Several time series foundation models, including Google's TimesFM (Das et al., 2024) and Amazon's Chronos (Ansari et al., 2024), have recently been released and applied across various fields such as economics (Carriero et al., 2024) and climate studies (Bodnar et al., 2024). These foundation models consistently outperform traditional statistical and deep learning models in these applications, highlighting their necessity and benefits for transportation research. In this study, we select Amazon's Chronos as our example because of its exceptional performance as a foundational model for time series analysis and the availability of multiple versions with different parameter configurations (Ansari et al., 2024).

The application of foundation models to transportation research is particularly relevant given the field's long history of technological innovation and the increasing complexity of modern transportation systems. A key innovation in this field is adaptive cruise control (ACC), which has evolved since the introduction of automated highway systems in the 1930s (Shladover, 1995; Xiao et al., 2017). ACC systems enhance driving comfort by automatically maintaining appropriate gaps between vehicles (Rahman et al., 2017), with studies indicating their potential to increase lane capacity at full market penetration (VanderWerf et al., 2001). Understanding ACC behavior through car-following models is essential for optimizing traffic flow efficiency and advancing autonomous vehicle technology (Goodall & Lan, 2020; Jurj et al., 2021). The European Commission Joint Research Centre's Open ACC dataset provides valuable vehicle trajectory data for studying these systems (Anesiadou et al., 2020).

With the development of autonomous vehicles, the need to better understand, simulate and predict the behaviors of ACC involved car-following situation become more and more important. This paper aims to examine the potential and capabilities of foundation models in the application of transportation data and the prediction of traffic flow conditions. While traditional mathematical car-following models can predict the condition of the followers based on the environment, foundation models can forecast the condition without receiving information of the current environment (i.e., without knowing the current speed of leading vehicle). This is a novel and unique approach, as these models have not yet been used in the transportation field and hold great potential. In this paper, we will use the time-series foundation models, Chronos to study car-following behavior with the Open ACC database. These models excel across various forecasting history lengths, prediction lengths, and time granularities at inference time (Ansari et al., 2024; Zhao et al., 2024). Their performance will be compared to the traditional mathematical model IDM and other time-series machine learning models. To the best of our knowledge, this is the first study to apply time series foundation models to the analysis of car-following behaviors.

## RELATED WORK

Car-following models, which describe the longitudinal interactions of adjacent vehicles, have been extensively studied for decades. The first model, proposed by Pipes (*29*) more than seventy years ago, laid the foundation for numerous subsequent mathematical models. Traditional mathematical base car-following models generally fall into five main categories: safe distance model, optimal velocity model, desired measure model (also known as desired goal model), stimulus-response model, and psycho-physiological models (*10*, *30–32*). As mentioned in the introduction, this paper is using IDM to predict the acceleration of the following vehicle as a baseline for performance comparison, where IDM was developed by Treiber et al (*4*). This accident-free, comprehensive model can respond to all traffic situations (*33*). It assumes each driver has specific goals, such as preferred following speed and headway, and assumes that drivers aim to reach both goals simultaneously. The IDM is the most common driver-based goal model and is widely applied with ACC data (7).

In addition to traditional models, the past decade has seen a surge in data-driven car-following models, driven by the availability of real-world driving data and advancements in machine learning. These models use AI techniques such as nonparametric regression, fully connected neural networks, recurrent neural networks, reinforcement learning, and other methods to predict driver behavior. They learn the relationships between various factors and driver behavior from the collected data (*10*). Recently, foundation models have gained more attention. A foundation model is trained on broad data, typically using self-supervision at scale, and can be adapted to various downstream tasks. Technologically, foundation models are not new; they are based on the techniques that have existed for decades: deep neural networks and self-supervised learning. However, the recent scale and scope of foundation models have expanded our understanding of the potential. For example, Yuan et al. (*34*) developed advanced foundation models for computer vision that can be easily adapted for various computer vision tasks, such as classification, retrieval, object detection, action recognition, etc. Moor et al. (*35*) proposed a new paradigm for medical AI: generalist medical AI (GMAI) based on foundation models, capable of performing a diverse set of tasks with minimal or no task-specific labeled data, such as assist versatile digital radiology in report drafting and assist surgical teams with procedures. Hong et al. (33) developed a universal remote sensing foundation model to focus on spectral data which can accommodate images with different characteristics and quality and captures spectrally sequential patterns. Foundation models also have great potential in handling time series data. González et al. (*36*) developed Foundation Auto-Encoders (FAE) model for anomaly detection in time-series data, and they have demonstrated that the model is able to capture and track the monthly variation of the data without previous evidence of it. Darlow et al. (35) developed a neural time series foundation model (DAM) which outperforms existing models at multivariate long-term universal time series forecasting.

However, none have explored the potential of time series foundation models in transportation research, which relies heavily on data such as signal timing, daily traffic flow changes, and car-following trajectories. Given that time series foundation models are relatively new, existing studies are limited. These pre-trained models offer significant advantages: they perform well on time series forecasts without specific dataset training and require only a small amount of data for fine-tuning to handle various downstream tasks. Therefore, this paper aims to explore the use of time series foundation models in analyzing car-following behaviors, investigate their potential in transportation research, and bridge the gap between modern machine learning methods and car-following studies. By combining ACC data with time series foundation models, this paper presents a novel and unique application in car-following models, aiming for improved performance and more accurate of describing car-following behaviors.

**METHODS**

In this section, we introduce the dataset and cleaning process, the IDM model, and the foundation models Chronos used in the experiments. The objective is to predict the acceleration of following vehicles $a^f$, as acceleration is a more direct and effective prediction of a vehicle's movement in the next time steps than speed, $v^f$ (*9*).

**Data Overview and Process**

The Casale experiment from the Open ACC dataset was selected for this paper. This data features two vehicles: the leader is driven manually, while the follower drives with ACC on at all times. Throughout the experiment, conducted on public roads in northern Italy (from Ispra to Casale Monferrato and back), the car-following order remained consistent. The dataset includes a full day of car-following testing with multiple records of vehicle trajectories, making it ideal for this paper's purposes. Data cleaning and processing was done by following Zhou et al (*37*). papers using Python. **Table 1** shows the overview of the Casale dataset.

**Table 1**: Overview of Casale dataset from Open ACC

| Data Collection | Sensor | Ublox 9 & On-Board Diagnostic (OBD) |
|---|---|---|
| | Accuracy | 0.3 m & 0.14 m/s |

| | Frequency | 10 Hz |
|---|---|---|
| Follower vehicle | Brand Model | Hyundai Ioniq Hybrid |
| | Year | 2018 |
| | Power System | Hybrid |
| | Dimension | Sedan |
| | ACC Setting | On |
| Test Sites | Accessibility | Public |
| | Road Type | Highway |
| | Free flow speed | 130 km/h |
| | Weather | Ideal |
| Experiment Setting | Duration | 10/27/2020 (one day) |
| | Speed Setting | no |
| | Headway Setting | no |
| | Drive Types | Naturalistic Drive |

After processing the raw data, the entire dataset is re-organized into different trajectories with assigned trajectory ID, there are 43 trajectories and in total 43,298 records identified from the original dataset. The acceleration and the position in the Frenet coordinate of leader and follower vehicles are calculated. The bump-to-bump distance and the speed difference of the two vehicles have also been calculated.

**Table 2** shows the statistical results of the dataset after the cleaning process, where $v^l$ and $a^l$ is the speed and acceleration of leader vehicles respectively, $v^f$ and $a^f$ are the speed and acceleration of follower vehicle respectively, and $g$ is the bump-to-bump distance between them (gap).

**Table 2:** Statistical results of cleaned Casale data

| *Variable* | *mean* | *std* | *min* | *max* | *unit* |
|---|---|---|---|---|---|
| $v^l$ | 32.6 | 5.7 | 0.2 | 40.1 | m/s |
| $a^l$ | 0.0 | 0.6 | -5.0 | 4.9 | m/s² |
| $v^f$ | 32.6 | 5.9 | 0.1 | 41.2 | m/s |
| $a^f$ | 0.0 | 0.6 | -4.7 | 4.9 | m/s² |
| $g$ | 42.7 | 13.7 | 1.6 | 119.6 | m |

**Intelligent Drive Model**

The IDM must be modified to accommodate local traffic conditions (e.g., the leading vehicle's condition) to accurately simulate car-following behaviors and predict the following vehicle's acceleration. The IDM used in this paper is developed from Treiber's model (*4*) and express in **Equation 1**:

$$a^f = a^{max}\left[1 - \left(\frac{v^f}{v^d}\right)^{c_0} - \left(\frac{d^*}{x^l - x^f}\right)^{c_1}\right] \tag{1}$$

with the following parameter $d^*$ in **Equation 2**:

$$d^* = d + c_2\sqrt{\frac{v^f}{v^d}} + v^f\tau + \frac{v^f(v^l - v^f)}{2\sqrt{a_{max}|b_d|}} \tag{2}$$

, where:

$a_{max}$ : the maximum acceleration, has range [0.73-5]
$v^d$ : desired speed, set to be free-flow speed 130 km/h
$c_0$ : model parameter, acceleration exponent, has range [0.2-20]
$d*$ : desired minimum gap, needs to be positive
$d$ : Desired standstill distance, has range [0-17]
$x^l - x^f$ : spacing between two vehicles
$c_1$ : model parameter, set to be 2
$c_2$ : model parameter
$\tau$ : minimum steady-state time gap, has range [1-2.5]
$b_d$ : comfortable deceleration, has range [2-9]

The value and the range of parameters comes from Treiber et al. (*4*), Milanés et al. (*7*), Kim et al. (*33*) and Souza et al. (*38*). The simulation is done using Python. The first 80% of the trajectories are used for training, and the remaining 20% of the trajectories for testing, and Root Mean Squared Error (RMSE) and Standard Deviation (Std) of the predicted $a^f$ is calculated as indicators of the model performance.

**Foundation Models**
Chronos, developed by Amazon, is a simple yet effective framework for pretrained probabilistic time series models. It scales and quantizes time series values into a fixed vocabulary, then uses cross-entropy loss to train existing transformer-based language models on these tokenized series (*18*). This model is based on the latest Large Language Models (LLMs) from the T5 (Text-to-Text Transfer Transformer) (*39, 40*) family, ranging from 20M to 710M parameters. It utilizes a large collection of publicly available datasets, supplemented by a synthetic dataset generated via Gaussian processes to enhance generalization. Chronos is based on the T5 architecture, and has five different sub-models that only difference in the vocabulary size. These models are chronos-t5-tiny (8 million parameters), chronos-t5-mini (20 million parameters), chronos-t5-small (46 million parameters), chronos-t5-base (200 million parameters), and chronos-t5-large (710 million parameters). In this paper, we use small, base and large models for the experiment.

*Objective function*
**Equation 3** shows the objective function of Chronos, it is trained to minimize the cross entropy between the distribution of the quantized ground truth label and the predicted distribution:

$$\ell(\theta) = -\sum_{h=1}^{H+1} \sum_{i=1}^{|v_{ts}|} 1_{(z_{C+h+1}=i)} \log p\theta\left(z_C + h + 1 = i \middle| z_{1:C+h}\right) \tag{3}$$

where the first $C$ is time step constitute the historical context, and the remaining $H$ represent the forecast horizon, $v_{ts}$ represents the two special tokens commonly used in language models, `PAD` (used to pad time series of different lengths to a fixed length for batch construction and to replace missing values) and `EOS` appended to the quantized and padded time series to denote the end of the sequence). $p\theta\left(z_C + h + 1 = i \middle| z_{1:C+h}\right)$ denotes the categorical distribution predicted by the model parameterized by $\theta$. The details of Chronos models theory refer to Ansari et al. paper (*18*). Chronos is probabilistic by design and multiple realizations of the future can be obtained by autoregressively sampling from the predicted distribution.

*Zero-shot forecasting*
Chronos has the ability to perform zero-shot forecasting, meaning it can generate forecasts for time series from unseen datasets without requiring explicit retraining on the target data. This out-of-the-pocket capability allows Chronos to achieve performance that is at least as good as advanced statistical models specifically tailored to the target dataset. This zero-shot forecasting ability significantly enhances its versatility and efficiency in handling diverse time series data. According to experiments by Ansari et al., Chronos models significantly outperform other pretrained models, including Moirai-1.0-R, Lag-Llama, LLMTime, ForecastPFN, and even GPT4TS (*18*). As a generalist time series forecaster, Chronos excels in zero-shot settings, surpassing commonly used local models and matching the performance of top task-specific deep learning models. Zero-shot Chronos models are used in the experiment as baseline cases.

*Covariates*

Chronos focuses on univariate time series forecasting, the most common real-world use case. However, practical forecasting tasks often involve additional information that must be considered. There are two types of covariates, Static covariates, which are covariates for each time series, and Dynamic covariates, which are covariates for each time stamps. Let's take an example of forecasting customers for a restaurant. Given the observed the daily numbers of customers of this month (30 days), forecast the daily number of customers of the next month (30 days). In this case, besides the daily number of customers, we also have covariates such `Category_of_food`, `Base_price`, `Weekday`, `Has_promotion`. Based on the example, `Category_of_food` is a static categorical covariate, and `Base_price` is a static numerical covariate. `Weekday` and `Has_promotion` are dynamic categorical covariates. Therefore, in the context of car-following model, in our experiments, the input variables: spacing between leader and follower vehicle (`Space_gap`), speed difference between two vehicles (`Speed_diff`), and speed of the following vehicle (`Speed_FAV`) are the dynamic covariates that need to be taking consider into the training process, to forecast our target value acceleration of following vehicle (`Acc_FAV`). These input variables are selected based on the review from Ma et al. (*9*) for papers that use time series machine learning methods such as LSTM for car-following predictions.

To train task-specific adaptors that incorporate covariates into the pretrained forecasting model, a regression machine learning model is needed. We have developed a novel approach: get the Chronos forecast of the `Acc_FAV` and fit the Light Gradient Boosting Machine (LightGBM) predicting residuals on the covariates ("Chronos + LightGBM") as a demonstration. LightGBM is a gradient boosting framework that uses tree-based learning algorithms. The processing of incorporating covariates into Chronos models are expressed by **Equations 4-10** below:

$$Output\left(a^f\right) = TF_{a^f}\left(a^f_{t-N}, \ldots a^f_{t-1}, a^f_t\right) \tag{4}$$

$$Output(\Delta x) = TF_{\Delta x}\left(\Delta x_{t-N}, \ldots, \Delta x_{t-1}, \Delta x_t\right) \tag{5}$$

$$Output\left(\Delta v\right) = TF_{\Delta v}\left(v_{t-N}, \ldots, v_{t-1}, v_t\right) \tag{6}$$

$$Output\left(v^f\right) = TF_{v^f}\left(v^f_{t-N}, \ldots, v^f_{t-1}, v^f_t\right) \tag{7}$$

$$e_{a^f} = V_{True} - Output_{a^f} \tag{8}$$

$$e_{forecast} = LightGBM\left(Output_{\Delta x}, Output_{\Delta v}, Output_{v^f}\right) \tag{9}$$

$$a^f_{forecast} = Output_{a^f} + Res_{forecast} \tag{10}$$

, where $TF$ representing Time Foundation Models, $a^f$ is the acceleration of the follower, $t$ is time step, $N$ is the context length, $\Delta x$ is the space gap between two vehicles, $\Delta v$ is the relative speed difference between the follower and the leader vehicles, $v^f$ is the speed of follower, $e$ is the residuals, and $V_{True}$ is the true values. In the experiment, we trained models both with and without covariates for each selected size of Chronos time series foundation models. "Without covariates" means the model forecasts based solely on historical data of $a^f$, while "with covariates" means the forecast considers input variables that mentioned above.

**EXPERIMENTS**

This section focuses on the experiment setting for training and testing Chronos foundation models. We conduct experiments for Chronos and run other time series machine learning models for comparison (i.e., serve as baselines).

**Baselines**

To evaluate the performance of Chronos models, we compared them with the following models:

- **IDM** (prediction model): It is a traditional mathematical car-following model to predict the acceleration of the following vehicle. It assumes each driver has specific goals, such as preferred following speed and headway, and assumes that drivers aim to reach both goals at the same time.
- **ETS** (exponential smoothing with trend and seasonality): This statistical model captures simple patterns in the data, such as trends and seasonality. In exponential smoothing with trend and seasonality, the error (E), trend (T), and seasonal (S) components are fixed and provided by the user (*41*)

- **DeepAR**: This deep learning model uses neural networks to capture complex patterns in the data. It is an autoregressive forecasting model based on a recurrent neural network. By training on a large number of related time series, this model can produce accurate probabilistic forecasts (*42*).
- **WaveNet**: This estimator employs the architecture proposed by Oord et al.in 2016 (*43*) with quantized targets. It utilizes a CNN architecture with dilated convolutions. Time series values are quantized into buckets, and the model is trained using cross-entropy loss.
- **TFT** (Temporal Fusion Transformer Model): This model combines LSTM with a transformer layer to predict the quantiles of future target values. It uses an attention-based architecture that provides high-performance multi-horizon forecasting along with interpretable insights into temporal dynamics (*44*).

**Time Foundation Model**

The time foundation model used in this experiment are Chronos with three different sizes: Chronos small with 46M parameters, Chronos base with 200M parameters, and Chronos large with 710M parameters.

**Setup**

In this paper, we conduct experiments using NVIDIA® L40 GPUs within a high-performance cluster (HPC) environment. This setup provided the computational power necessary for training complex models efficiently. The dataset is divided into training and testing, where the first 80% for the trajectories are for training and the remaining 20% are held-out for testing.

**Evaluation Metric**

In order to compare the performance across different models, we calculate the RMSE between the predicted/forecasted and actual value of $a^f$. We can more accurately estimate performance using backtesting, which evaluates multiple forecast horizons from the same time series. As shown in **Figure 1**, this method measures forecast accuracy using the last prediction length time steps of each validation split as a hold-out set (marked in yellow). Each of the boxes representing 1 second, the greed boxes representing historical or true value, the yellow boxes representing the predicted/forecasted value. In the first step, the first 3 seconds are predicted by the historical 6 seconds. In the second step, it includes the true value of the future 3 seconds and calculates the error between the prediction from the first step and the true value, then predict/forecast the next 3 seconds. Multi-window backtesting generally provides a more accurate estimation of forecast quality on unseen data.

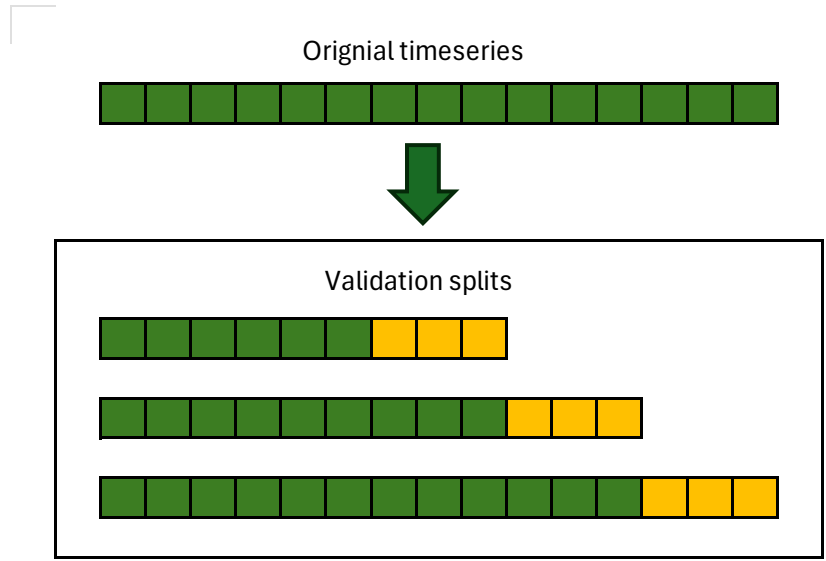

**Figure 1:** Backtesting using multiple windows

The mean RMSE is calculated as shown in **Equation 11** and **12**.

$$RMSE = \sum_{t-1}^{T} \frac{\sqrt{\left(a^f_{\;i} - \widehat{a^f_{\;i}}\right)^2}}{T} \tag{11}$$

$$T = \frac{len\;(\;test\;set)}{len(\;Time\;Windows)} \tag{12}$$

, where *len* is the length of the dataset.

**Fine-Tuning**

In addition to using the zero-shot Chronos model, we also fine-tune the models. Fine-tuning involves adapting a pre-trained model to a specific downstream task using a smaller, task-related dataset. This process leverages the general knowledge acquired during pre-training and customizes the model for particular applications, such as forecasting car-following behavior. Fine-tuning allows for quicker training and often better performance by utilizing features learned during pre-training. In contrast, training from scratch starts with a model with randomly initialized weights, requiring more time, computational resources, and a higher risk of overfitting. In this paper, Chronos is fine-tuned specifically to forecast $a^f$ with the training dataset.

**RESULTS**

In this section, we aim to compare the performance of the time series foundation model with several baseline models. Also, we justify ensemble learning (*45*) to incorporate covariates and show the model's performance with varying sizes. Finally, we present visualizations of the continuous forecasted acceleration of following vehicles alongside the true values.

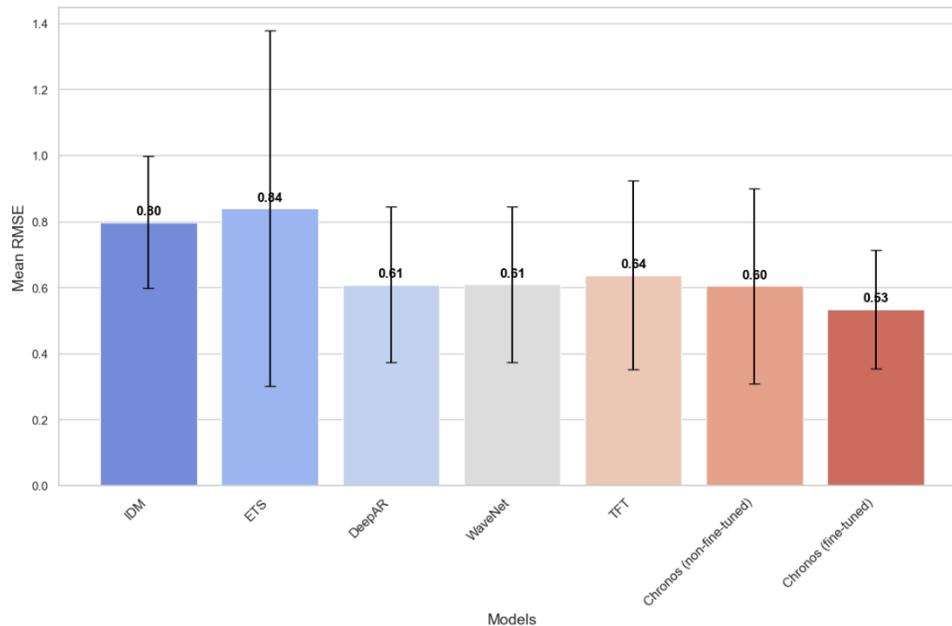

**Figure 2**: RMSE for forecasting the acceleration of following vehicles.

In **Figure 2**, the numbers on top of each bar represent the mean RMSE achieved by each model across multiple time windows, while the black line indicates the standard deviation across these windows. The fine-tuned Chronos models, which incorporate covariates, achieved the lowest mean and standard deviation of RMSE. Among all the baseline models, the statistical model EST performed the worst, with a mean RMSE of 0.84. The car-following model IDM performed slightly better than EST, with a mean RMSE of 0.80. Although not as good as the fine-tuned version, the non-fine-tuned Chronos model achieved similar performance to other deep learning models such as DeepAR, WaveNet, and TFT, with a mean RMSE of 0.60. This indicates that the foundation model can be effectively applied out-of-the-box, even without further training. These results highlight the high potential of the foundation model in forecasting car-following conditions. With fine-tuning of training dataset, Chronos has the best performance over all the models with mean RMSE of 0.53, this is 33.75% of improvement over IDM.

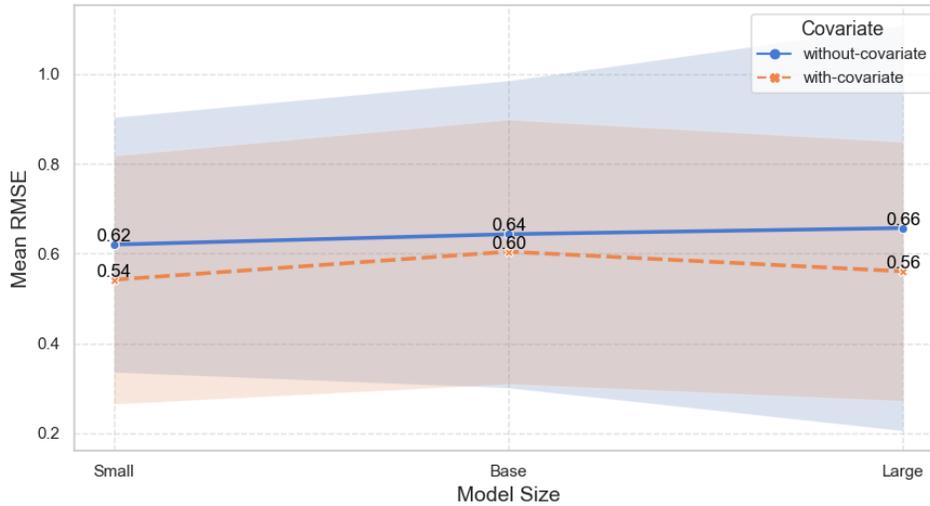

**Figure 3**: Comparison of the performance with and without incorporating covariates and the differences observed using various sizes of Chronos models.

**Figure 3** illustrates the effectiveness of incorporating covariates and the differences observed using various sizes of Chronos models. From the figure, it is evident that the small, base, and large-sized Chronos models achieved similar RMSE, with the small model having the lowest value. This result indicates that the small size is sufficient to meet the requirements, which is advantageous as it is more time and computationally efficient. For all three model sizes, the performance with covariates is better than without covariates. This demonstrates that incorporating covariates can enhance the performance of the time foundation models when applied to vehicle trajectory data.

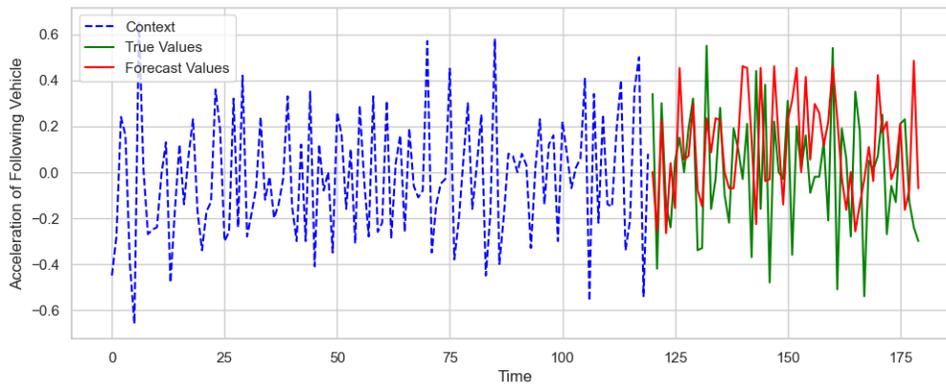

**Figure 4:** Visualization of the forecast versus actual acceleration of the following vehicle within one time window.

**Figure 4** shows the acceleration data of the following vehicle over time, comparing the context (dashed blue line), the true values (green line), and the forecast values from the fine-tuned Chronos model with covariates (red line). Initially, up to around the 120th time unit, the context data illustrates the historical acceleration patterns of the vehicle. Beyond this point, the true and forecasted acceleration values are plotted within one time window. The close alignment between the green and red lines at the beginning of the forecasting period suggests that the model accurately captures the acceleration trends of the following vehicle in the short term. However, as time progresses, some discrepancies between the forecasted and actual values emerge, indicating that long-term forecasting remains challenging. Overall, the close alignment between the green and red lines demonstrates the model's ability to capture the acceleration trends of the following vehicle.

## DISCUSSION & CONCLUSIONS

As shown in the result section, comprising traditional mathematical models and deep learning methods, Chronos significantly outperform other methods on the trajectory datasets for forecasting acceleration of following vehicle. We demonstrate that Chronos can achieve similar performance with other deep learning models at zero-shot and has the best performance after fine-tuning with training dataset. Despite the presence of errors, as indicated by the 0.53 RMSE, time series foundation models demonstrate great potential in handling transportation time series data. We acknowledge that this paper applied only one dataset and one type of time series foundation model. However, the main contribution lies in our exploration of applying time series foundation models to transportation data, specifically using car-following trajectory data as an example. This study highlights the promising performance of these models in forecasting vehicle conditions. Moreover, this study used one traditional mathematical model as a baseline. Future research could include additional traditional mathematical models, such as the optimized velocity model, to provide a more comprehensive comparison. In future work on car-following behavior analysis, we plan to incorporate additional time series foundation models and apply them to other available datasets, such as the Waymo Open dataset and the Vanderbilt ACC dataset. Including more datasets is expected to enhance the fine-tuning of time series forecasting models and enable cross-dataset evaluation, where an entire dataset can be reserved for evaluation.

## ACKNOWLEDGMENTS
The authors want to thank Open ACC database for providing car-following experiment vehicle trajectory data. The authors also want to thank Professor Alexandra Kondyli and Professor Chris Depcik for proving suggestions and comments on the concepts. Lastly, the authors want to thank ChatGPT for proofreading and gramma checking.

## AUTHOR CONTRIBUTIONS
The authors confirm contribution to the paper as follows: study conception and design: L. Zeng, R. Yan; data collection: L. Zeng; analysis and interpretation of results: L. Zeng, R. Yan; draft manuscript preparation: L. Zeng, R. Yan. All authors reviewed the results and approved the final version of the manuscript.